%% file: main.tex
\documentclass[10pt,twocolumn,letterpaper]{article}

\usepackage[pagenumbers]{iccv} 

\usepackage[accsupp]{axessibility}
\usepackage{booktabs}
\usepackage{multirow}
\usepackage{siunitx}
\usepackage{tabularray}
\usepackage{graphicx}
\usepackage{array}
\usepackage{tikz}
\usetikzlibrary{patterns} 
\usepackage{tabularray}

\usepackage{stfloats}
\usepackage{relsize}

\usepackage{colortbl} 

\newcommand{\vlcolor}{black}
\newcommand{\vl}[1]{\textcolor{\vlcolor}{#1}}


\definecolor{iccvblue}{rgb}{0.21,0.49,0.74}
\definecolor{myblue}{HTML}{0B30EA}
\usepackage[pagebackref,breaklinks,colorlinks,allcolors=iccvblue]{hyperref}

\def\paperID{5049} 
\def\confName{ICCV}
\def\confYear{2025}

\title{Egocentric Action-aware Inertial Localization in Point Clouds with Vision-Language Guidance}

\author{Mingfang Zhang$^1$,~~Ryo Yonetani$^2$,~~Yifei Huang$^1$,~~Liangyang Ouyang$^1$,~~Ruicong Liu$^1$,~~Yoichi Sato$^1$\\
$^1$The University of Tokyo,~~$^2$CyberAgent AI Lab\\
{\tt\small \{mfzhang,hyf,oyly,lruicong,ysato\}@iis.u-tokyo.ac.jp,~~yonetani\_ryo@cyberagent.co.jp}
}

\begin{document}
\maketitle
\input{./sec/0_abstract}

\input{./sec/1_intro}
\input{./sec/2_related_works}

\input{./sec/3_problem_setting}
\input{./sec/4_method}
\input{./sec/5_experiments}
\input{./sec/6_conclusion}

{
    \small
    \bibliographystyle{ieeenat_fullname}
    \bibliography{main}
}

\input{sec/X_suppl}

\end{document}

%% file: sec/0_abstract.tex
\begin{abstract}

This paper presents a novel inertial localization framework named Egocentric Action-aware Inertial Localization (EAIL), which leverages egocentric action cues from head-mounted IMU signals to localize the target individual within a 3D point cloud. 
Human inertial localization is challenging due to IMU sensor noise that causes trajectory drift over time. The diversity of human actions further complicates IMU signal processing by introducing various motion patterns.
Nevertheless, we observe that some actions captured by the head-mounted IMU correlate with spatial environmental structures
(e.g., bending down to look inside an oven, washing dishes next to a sink), thereby serving as spatial anchors to compensate for the localization drift.
The proposed EAIL framework learns such correlations via hierarchical multi-modal alignment \vl{with vision-language guidance}. By assuming that the 3D point cloud of the environment is available, it contrastively learns modality encoders that align short-term egocentric action cues in IMU signals with local environmental features in the point cloud. \vl{The learning process is enhanced using concurrently collected vision and language signals to improve multimodal alignment.} The learned encoders are then used in reasoning the IMU data and the point cloud over time and space to perform inertial localization. Interestingly, these encoders can further be utilized to recognize the corresponding sequence of actions as a by-product.
Extensive experiments demonstrate the effectiveness of the proposed framework over state-of-the-art inertial localization and inertial action recognition baselines. Project page: \small{\url{https://github.com/mf-zhang/Ego-Inertial-Localization}}.

\end{abstract}

%% file: sec/1_intro.tex
\vspace{-1em}

\section{Introduction}
\label{sec:intro}

\begin{figure}[t]
  \centering
   \includegraphics[width=1.\linewidth]{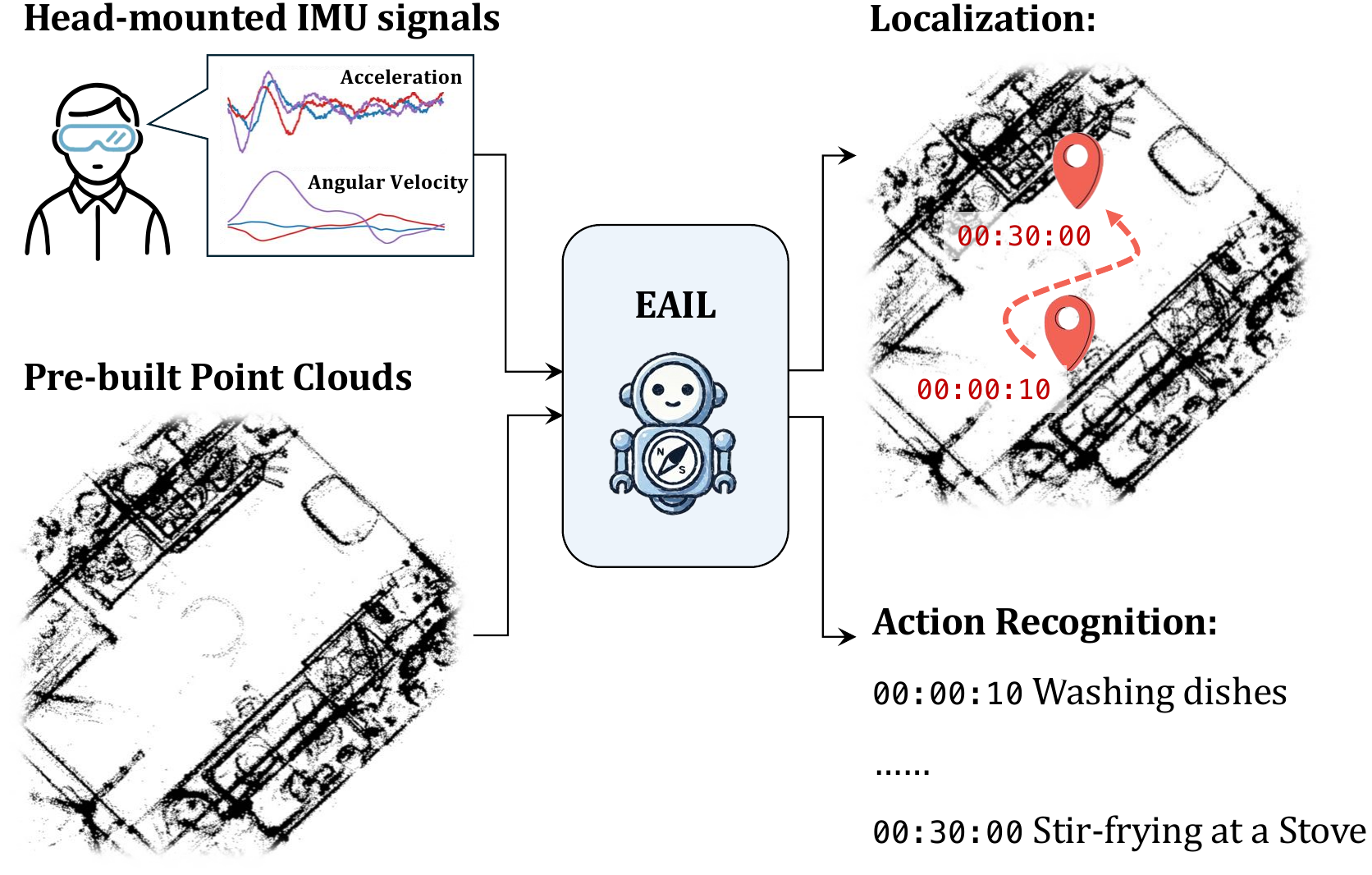}
   \caption{\textbf{Egocentric Action-aware Inertial Localization \textsmaller{(EAIL)}}. Our framework leverages egocentric action cues obtained from the head-mounted IMU to perform inertial localization in the environmental 3D point cloud. The corresponding sequence of actions can also be recognized as a by-product.}
   \label{fig:teasor}
   \vspace{-1em}
\end{figure}

Inertial localization has emerged as a pivotal technology for human tracking across a wide range of various applications, including personal navigation \cite{herath2020ronin} and augmented reality~\cite{sarlin2022lamar}. Inertial measurement units (IMUs) are commonly embedded in wearable devices such as smartphones, smart glasses \cite{engel2023project}, and headsets \cite{park2021review}. When affixed to human bodies, IMUs can capture acceleration and angular velocity to record 3D human movements. Compared to vision-based localization methods \cite{lowe2004distinctive, kendall2015posenet}, inertial localization enables user tracking in an energy-efficient and privacy-preserving manner.

Despite the advantages, human inertial localization remains highly challenging primarily due to two reasons. 
The first one is \emph{the trajectory drift caused by IMU noise}. IMU sensor noise causes small uncertainties in measurements that accumulate over time, which can ultimately lead to significant trajectory drift \cite{titterton2004strapdown}.
Conventional step detection methods \cite{brajdic2013walk, ying2007automatic} struggle to generalize to the noise of irregular movements, while recent data-driven approaches \cite{herath2020ronin, zeinali2024imunet, sun2021idol, liu2020tlio}, which predict velocity from IMU signals, can suffer from accumulated estimation errors.

The other challenge lies in \emph{the complexity of human actions}. Human body-mounted IMUs capture signals not only from displacement-related movements like walking and stopping, but also from action-induced motions that do not involve actual positional changes, such as head-swinging during cooking and cleaning. These additional motion signals can complicate IMU signal processing and make inertial localization further difficult. 
Existing datasets \cite{yan2018ridi, herath2020ronin, chen2018oxiod} and approaches \cite{herath2022neural,yan2018ridi} focus mostly on human walking scenarios, limiting the ability of state-of-the-art inertial localization methods to handle real-world human action variability. 

Nevertheless, we argue that human actions can rather act as a salient locational cue to mitigate the trajectory drift challenge if properly taken into account. For example, washing dishes often takes place near a sink, while frying often occurs near a stove. Our key insight is that extracting such actions from IMU data can help to deduce the spatial regions where they are likely to occur.


In this work, we present a novel framework named \emph{Egocentric Action-aware Inertial Localization (EAIL; see also \cref{fig:teasor})}. While assuming that the 3D point cloud of the environment is readily available using off-the-shelf 3D scanners, this framework leverages egocentric action cues extracted using head-mounted IMUs to localize the target individual in the environmental point cloud. 

The key technical contribution is a hierarchical modality alignment technique that effectively extracts the correlation between the action cues and the point cloud \vl{with vision-language guidance}.
Specifically, we first learn the encoders for short-term IMU signals and local point clouds to be aligned with the corresponding egocentric point-of-view images and textual descriptions of observed actions in a contrastive fashion. \vl{By incorporating guidance from pretrained vision-language models, we achieve improved multimodal feature alignment.} The learned IMU and point-cloud encoders are then invoked in spatio-temporal reasoning modules to jointly predict possible actions occurring at each moment and the locations in the global point cloud where the predicted actions can be observed. While our primary goal is the localization in the latter part, the predicted locations can also improve inertial action recognition as a beneficial by-product.

Extensive evaluations on the EgoExo4D dataset \cite{grauman2024ego} validate that our framework achieves state-of-the-art performance in both inertial localization and inertial action recognition compared to \cite{herath2022neural, zeinali2024imunet, moon2023imu2clip, zhang2024masked}. 
In summary, our main contributions are as follows:

\begin{itemize}

\item We introduce EAIL, a novel inertial localization framework that leverages egocentric action cues from head-mounted IMU signals to localize target individuals within a 3D point cloud.

\item We develop a hierarchical 
modality alignment
technique that learns the correlation between the inertial egocentric action cues and the environmental point cloud to perform inertial localization as our primary goal, and action recognition as a by-product. 

\item Extensive evaluations demonstrate that our framework achieves state-of-the-art results in both inertial localization and inertial action recognition in diverse settings.

\end{itemize}

%% file: sec/2_related_works.tex
\section{Related Work}
\label{sec:related_works}

\subsection{Human Inertial Localization}

Human inertial localization has been an active research area. Leveraging both visual and inertial data, accurate localization can be achieved by perceiving and modeling the surrounding environment \cite{mourikis2007multi, leutenegger2015keyframe, qin2018vins}. Visual-inertial fusion can also simultaneously perform localization and body poses estimation \cite{liu2024egohdm, yi2023egolocate}. However, visual data availability is not guaranteed at all times, especially in human-centric applications where privacy concerns are essential.

Traditional inertial localization methods like Pedestrian Dead Reckoning (PDR) \cite{shu2015last, brajdic2013walk, ying2007automatic} estimate trajectories by detecting steps and estimating step length and heading. While effective in structured environments, these approaches struggle with noise from irregular human motion. Data-driven inertial navigation methods \cite{chen2018ionet, zeinali2024imunet, herath2020ronin} mitigate sensor noise by learning to estimate velocity from IMU signals. However, when they cumulate the estimated velocity to predict human locations, the estimation errors can lead to significant trajectory drift.

To mitigate the drift problem, external signals such as GPS \cite{kim2021deep}, Wi-Fi \cite{herath2021fusion}, Bluetooth \cite{li2015bluetooth}, user history \cite{yonetani2024retailopt}, and activities \cite{hardegger2012actionslam, zhang2022positioning}  have been used to provide anchor points for human localization. However, these solutions introduce dependencies on specific infrastructures or hand-defined landmarks. More recently, NILoc \cite{herath2022neural} takes a different approach by directly predicting user locations without explicit velocity integration, but it requires scene-specific training and lacks adaptability across diverse environments.

Finally, beyond these methodological challenges, existing datasets \cite{herath2020ronin, chen2018oxiod} and approaches \cite{herath2022neural,yan2018ridi} are primarily centered around displacement-centric activities such as walking and stopping. This focus overlooks a broader spectrum of complex human actions, thus limiting the models' abilities to manage real-world human motion variability. 
In contrast, our framework addresses the trajectory drift problem by leveraging egocentric action cues from IMU signals, to localize the user within a point cloud.

\subsection{Egocentric Multimodal Alignment}

Recent advancements in head-mounted devices \cite{engel2023project, sarlin2022lamar} have generated significant interest in understanding human activities from an egocentric perspective. Several large-scale egocentric datasets \cite{grauman2022ego4d, grauman2024ego, damen2018scaling, huang2024egoexolearn} have been introduced, offering rich multimodal data including video, gaze, audio, language, and IMU signals. These datasets provide a foundation for developing multimodal human understanding methods \cite{gong2023mmg, moon2023imu2clip, liu2024single, lin2025simhand}.

Aligning egocentric visual data with textual action descriptions enhances comprehension and broadens application possibilities. Large-scale video-language pretraining, specifically tailored to egocentric data, has been pioneered by works like EgoVLP \cite{kevin2022egovlp, pramanick2023egovlpv2} and HierVL~\cite{Ashutosh_2023_CVPR}. In addition to the human language narrations, LaViLa~\cite{zhao2023learning} refines these narrations with Large Language Models (LLMs). Recap~\cite{islam2024video} and Vinci~\cite{huang2024vinci} align fused languages for long videos. EMBED~\cite{dou2024unlocking} and EgoInstructor~\cite{xu2024retrieval} use rules or retrieval models to add additional training data.
Subsequent works further incorporate additional egocentric-specific cues such as hand~\cite{zhang2023helping} and audio~\cite{chen2024soundingactions}. These approaches have demonstrated that egocentric-specific pretraining yields transferable representations for various downstream tasks, such as question answering \cite{fan2019egovqa}, robot manipulation \cite{pei2025modeling}, and narration-based segmentation \cite{shen2024learning}.

Aligning vision and language embeddings with 3D environments \cite{xue2023ulip, liu2024vsrd, zhang2023structural} makes open-vocabulary affordance detection possible. Affordance denotes potential actions in the environment that an agent can perform to an object or in an area. \cite{delitzas2024scenefun3d}~provides accurate interaction annotations for real-world 3D indoor scenes. Works like \cite{takmaz2023openmask3d, gu2024conceptgraphs, lu2023ovir} utilize free-form queries describing geometry and affordances, leveraging the pretrained CLIP \cite{radford2021learning} model, to retrieve 3D instances. These studies show that a strong correlation between actions and environments can be learned through multimodal alignment.

Inspired by these advances, our framework introduces learning the multimodal alignment between the human action cues and the environmental structures, to serve as natural anchors in the scene to predict human trajectories.

%% file: sec/3_problem_setting.tex
\section{Problem Setting}
\label{sec:problem_setting}

\begin{figure*}[t]
  \centering
   \includegraphics[width=\linewidth]{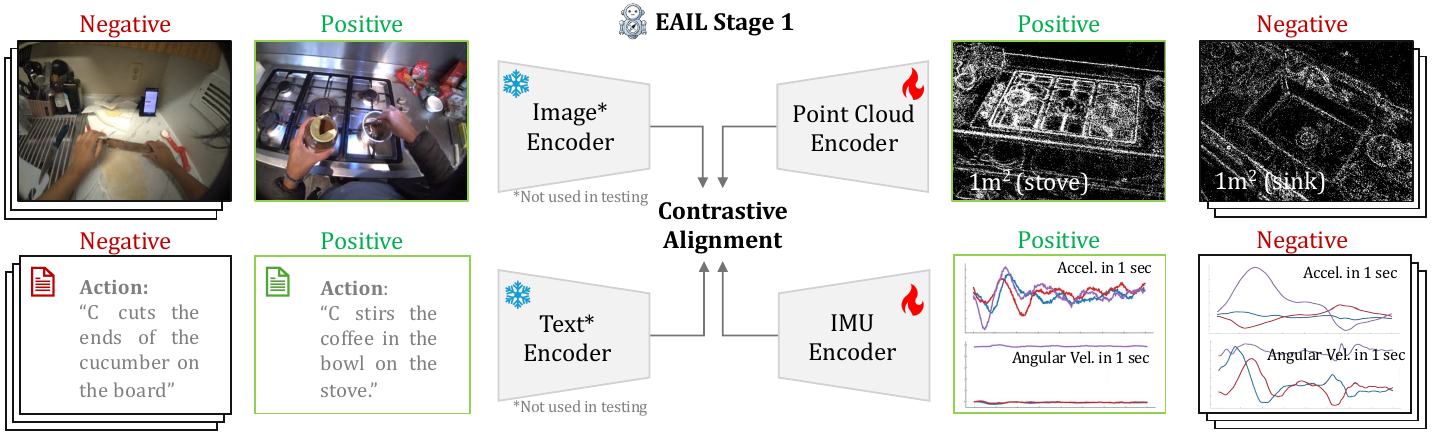}

   \caption{\textbf{Short-Term Action-Location Alignment.} In this first stage, our objective is to train a point cloud encoder and an IMU encoder using contrastive learning. A positive sample set consists of simultaneous multimodal data samples at the human's location. Negative samples are generated by randomly sampling data from other times and locations.}
   \label{fig:stage1}
   \vspace{-1em}
\end{figure*}


We address the task of inertial localization using head-mounted devices equipped with IMUs.
We assume that the 3D point cloud of the environment is available using consumer-grade 3D scanner apps~\cite{labbe2019rtab, scaniverse}. Our primary goal is to predict the sequence of the device user's locations in the point cloud.


Formally, let $\boldsymbol{\mathcal{M}} = [\textbf{M}_1, \dots, \textbf{M}_T]$ represent the sequence of IMU data over $T$ seconds, where each $\textbf{M}_t$ corresponds to the IMU signals captured within a single second. These signals comprise a sequence of 6-DoF sensor readings, specifically consisting of 3-axis accelerations and 3-axis angular velocities. We also denote the point cloud of the environment by $\boldsymbol{\mathcal{P}}$. The sequence of locations of the target individual on the ground,  $\boldsymbol{\mathcal{Z}} = [\textbf{z}_1,\dots,\textbf{z}_T],\; \textbf{z}_t\in\mathbb{R}^2$, is represented in the same coordinate system as that of~$\boldsymbol{\mathcal{P}}$. Our inertial localization method will learn a mapping $ (\boldsymbol{\mathcal{M}}, \boldsymbol{\mathcal{P}}) \longmapsto \boldsymbol{\mathcal{Z}}$.

\looseness=-1
Note that this problem setup is different from existing inertial navigation (\textit{e.g.}, \cite{herath2020ronin}) and inertial localization~\cite{herath2022neural}. Inertial navigation methods just predict velocity and thus require a known initial position to perform localization, making them prone to cumulative drift. Existing inertial localization~\cite{herath2022neural} operates in a scene-specific manner, training and evaluating models separately for each environment while disregarding explicit environmental structure. In contrast, our approach incorporates the 3D point cloud $\boldsymbol{\mathcal{P}}$, enabling localization without requiring environment-specific training.

%% file: sec/4_method.tex
\section{The EAIL Framework}
\label{sec:method}

\begin{figure*}[t]
  \centering
   \includegraphics[width=\linewidth]{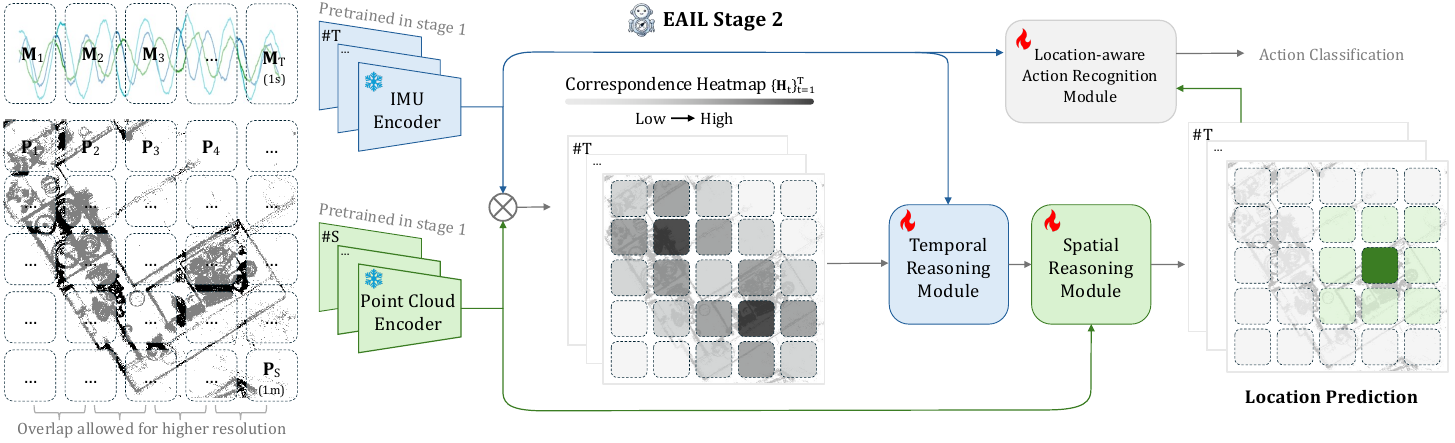}

   \caption{\textbf{Sequential Motion Localization.} In this second stage, we generate a sequence of the user's locations and actions over $T$ seconds using a series of IMU signals ($\{\textbf{M}_{t}\}_{t=1}^{T}$) alongside the point cloud of the entire scene. This point cloud is divided into $S$ local segments ($\{\textbf{P}_{s}\}_{s=1}^{S}$). The IMU encoder and the point cloud encoder are frozen networks pre-trained in Stage~1 for efficient spatial-temporal reasoning in large-scale point clouds.}
   \label{fig:stage2}
\end{figure*}

The proposed EAIL framework leverages action cues from head-mounted IMU signals to localize target individuals within the 3D point cloud. It proceeds in two stages:

\begin{itemize} 
    \item \textbf{Stage~1: Short-term action-location alignment} (\cref{fig:stage1}). 
    We first train an IMU encoder for processing short-term IMU signals (\eg, spanning 1 second) as well as a point-cloud encoder for local point clouds (\eg, 1 m$^2$). Here we propose a novel modality alignment technique that aligns the features of those IMU and point cloud inputs with the corresponding egocentric point-of-view images and text annotations of the actions taken by the target individual, which makes the trained encoders `action-aware'. Note that these egocentric images and text annotations are necessary only in this training stage.
    \item \textbf{Stage~2: Sequential motion localization} (\cref{fig:stage2}). 
    We then learn spatiotemporal reasoning modules that invoke the IMU/point-cloud encoders to predict the sequence of target individual's locations (and the sequence of their actions as a bonus; see \cref{subsec:stage2}). 
    
\end{itemize}



\subsection{\textbf{Stage~1: Short-term Action-Location Alignment}}

In the first stage, we aim to train a short-term IMU signal encoder and a local point cloud encoder. 
In the training, we focus on capturing action semantics to establish a meaningful correspondence between the two modalities. 
These correlations, such as the association between dish-washing motion and the structure of a sink, can serve as natural anchors to enhance downstream localization accuracy. 

To achieve this goal, we employ multimodal contrastive learning. 
Egocentric datasets \cite{grauman2024ego, grauman2022ego4d, damen2018scaling} usually include vision and text data alongside IMU data. We conduct contrastive alignment across four modalities: images, textual action descriptions, IMU signals, and point clouds, as shown in \cref{fig:stage1}.
In this way, we can utilize the robust capabilities of pre-trained and pre-aligned vision-language models \cite{mu2022slip, radford2021learning, chen2024internvl}.
The action semantics embedded in vision and text modalities aid to guide the training of the IMU and the point cloud encoders.
Note that the image and the language modalities 
are not required during testing.

Specifically, for each 1-second time segment, we extract a synchronized set of inputs: an ego-view image frame $\textbf{I}_t$, an action caption $\textbf{L}_t$, an IMU segment $\textbf{M}_{t}$, and a local $1 m^2$ region of point cloud ${\textbf{P}}_{t}$. 
We utilize off-the-shelf pre-trained vision-language encoders $E_I$ and $E_L$ to encode the image and the text data, while we train the IMU and the point cloud encoders, $E_M$ and $E_P$.
Then, we apply a pairwise contrastive loss $L_c$ across the generated features $\textbf{F}^I$, $\textbf{F}^L$, $\textbf{F}^M$, and $\textbf{F}^P$ derived from the four modalities:

\begin{equation}
\begin{split}
    L_{stage1}= \alpha L_c(\textbf{F}^I,\textbf{F}^M) + \beta L_c(\textbf{F}^I,\textbf{F}^P) \\
    + \theta L_c(\textbf{F}^L,\textbf{F}^M) + \delta L_c(\textbf{F}^L,\textbf{F}^P)
    + \gamma
    L_c(\textbf{F}^M,\textbf{F}^P).
\end{split}
\label{eq:stage1_loss}
\end{equation}

Through this process, our IMU and point cloud encoders are trained to generate features in a shared embedding space where multimodal features are connected through human action semantics.
This enables the retrieval of local point cloud segments with short-term IMU signals. It can provide a preliminary estimate of the user's location with IMU signals. However, given that similar motion patterns might occur in multiple locations, we introduce the second stage of our framework to address these ambiguities.

\subsection{Stage~2: Sequential Motion Localization}
\label{subsec:stage2}

In Stage~2, we build on the pre-trained short-term IMU encoder and the local point cloud encoder, and invoke them for processing sequential motion captured over $T$ seconds of IMU signals $\boldsymbol{\mathcal{M}}$ and the global point cloud $\boldsymbol{\mathcal{P}}$ of the entire scene to predict the user's trajectory $\boldsymbol{\mathcal{Z}}$.

\subsubsection{Spatiotemporal reasoning for trajectory prediction}

Predicting user trajectories within large-scale point clouds requires fine-grained temporal and spatial reasoning. 
To achieve this, as shown in \cref{fig:stage2}, we decompose the $T$-second IMU signals $\boldsymbol{\mathcal{M}}$ into short sub-sequences, $\{\textbf{M}_{t}\}_{t=1}^{T}$. Concurrently, the global point cloud is uniformly partitioned into $S$ local segments, denoted as $\{\textbf{P}_{s}\}_{s=1}^{S}$. Then, we apply the frozen \( E_M \) to extract IMU features \(\{\textbf{F}^M_{t}\}_{t=1}^{T}\). In parallel, the frozen \( E_P \) processes local point cloud segments to produce point cloud features \(\{\textbf{F}^P_{s}\}_{s=1}^{S}\).

Given that IMU features \(\textbf{F}^M\) and point cloud features \(\textbf{F}^P\) are aligned due to the Stage~1 contrastive learning process, we can generate a sequence of \textit{correspondence heatmaps} \(\{\textbf{H}_{t}\}_{t=1}^{T}\) by calculating the similarity between these two feature sets. Regions with a high score on these heatmaps indicate a high likelihood that human motion occurs in these spatial regions. For instance, ``washing'' would align closely with a ``sink'' region, while ``walking'' covers ``open areas''.

Building on this, we design a Temporal Reasoning Module and a Spatial Reasoning Module. The temporal reasoning module takes the correspondence heatmaps \(\{\textbf{H}_{t}\}_{t=1}^{T}\) and IMU features \(\{\textbf{F}^M_{t}\}_{t=1}^{T}\) as inputs and employs a 3D convolutional network to reason across the temporal dimension, generating refined features $\textbf{F}^R$. Next, $\textbf{F}^R$ is inputted, along with the point cloud features \(\{\textbf{F}^P_{s}\}_{s=1}^{S}\), into the spatial reasoning module. This module uses another dilated 3D convolutional network to reason in the 2D spatial dimension. The final output is the predicted user trajectory \(\boldsymbol{\mathcal{Z}}\).

For the trajectory \(\boldsymbol{\mathcal{Z}}\), we frame the prediction as an \mbox{$S$-class} classification problem. The goal is to find a sequence of point cloud segments that are nearest to the actual user trajectory. We use cross-entropy loss for training:
\begin{equation}
    L_{traj} = -\sum_{t=1}^{T} \sum_{s=1}^{S} \textbf{y}_{t,s} \log(\hat{\textbf{y}}_{t,s}),
\end{equation}
\noindent
where \(\textbf{y}_{t,s}\) is the true label indicating the user's nearest point cloud segment and \(\hat{\textbf{y}}_{t,s}\) is the predicted probability for time step \(t\) and segment $P_s$. 

\subsubsection{Location-aware action recognition}

In the Location-aware Action Recognition Module, we leverage the predicted user locations to enhance inertial action recognition. Specifically, for each time step $t$, we take all the predicted location probabilities \(\hat{\textbf{y}}_{t,s}\) in a scene, collectively represented by heatmaps $\{\textbf{H}'_{t}\}_{t=1}^T$, and use them as a spatial attention over the point cloud features $\{\textbf{F}^P_{s}\}_{s=1}^{S}$. 
We then blend these spatial features with IMU features $\{\textbf{F}^M_{t}\}_{t=1}^T$ through addition. Finally, a multi-layer perceptron maps the fused representation to action likelihood. The training is supervised by a cross-entropy loss:
\begin{equation}
    L_{action} = -\sum_{t=1}^{T} \sum_{c \in \mathcal{C}} \boldsymbol{y}_{t,c} \log(\hat{\boldsymbol{y}}_{t,c}),
\end{equation}
\noindent
where \( \boldsymbol{y}_{t,c} \) represents the true label for the action class \( c \) at time \( t \), and \(\hat{\boldsymbol{y}}_{t,c}\) is the predicted probability.

In summary, our whole model in Stage~2 is supervised by $L_{stage2}=L_{traj}+L_{action}$.

%% file: sec/5_experiments.tex
\section{Experiments}
\label{sec:experiments}

\subsection{Experimental Setup}

\paragraph{Dataset}

We use the EgoExo4D dataset \cite{grauman2024ego} that provides synchronized egocentric video and IMU signals recorded with Aria glasses \cite{engel2023project}. Prior to data collection, environment point clouds were built, and action captions were annotated afterward. For this paper, we used the cooking activities subset, which includes 173 participants across 60 kitchens, totaling 564.13 hours of recordings. The activity area for these cooking activities averaged around 2.8 meters per side, with the largest spanning 6.15 meters. Each local 1 $m^2$ point cloud is sub-sampled to contain 8192 points, and the IMU signals are recorded at an 800 Hz sample rate. The dataset is labeled with 35 distinct action classes for classification tasks. We divided the data into training, test-seen, and test-unseen sets. The test-unseen set consists exclusively of environments and participants not present in the training set. 

\paragraph{Implementation Details}
In Stage~1, we employed an enhanced version of the CLIP model \cite{mu2022slip} as the vision-language encoder. Specifically, we used ViT-Base \cite{dosovitskiy2020image} as the image encoder and the Base Text Transformer \cite{vaswani2017attention} from CLIP for text encoding. For the point cloud encoder, we adopted PointNet++ \cite{qi2017pointnet++}, and the IMU encoder was built upon ResNet18-1D \cite{he2016deep}. The IMU signals are pre-processed following \cite{herath2020ronin} and downsampled to 400 Hz. In Stage~2, our model operated with a sequence length \( T = 10 \), uniformly partitioning the point cloud into \( S = 20 \times 20 = 400 \) local segments. We trained Stage~1 for 250 epochs and Stage~2 for 100 epochs, using a batch size of 64, a learning rate of \( 10^{-3} \), and the AdamW optimizer. The loss parameters were set as follows: $\alpha=0.1;\ \beta,\theta,\delta,\gamma=1$.

\paragraph{Evaluation Metrics}
For the localization task, we report the success rate (\%) at error distance thresholds of 0.2 m, 0.4 m, and 0.6 m following NILoc \cite{herath2022neural}. Additionally, we include the Relative Score (RS) metric, which measures the proportion of heatmap positions with lower confidence than the ground truth location. This metric indicates the relative ranking of the prediction at the target point. For action classification, we evaluate performance using top-1 and top-5 accuracy metrics. These scores are assessed under two setups: ``seen rooms'' where the localization is performed in the environments present in the training dataset and ``unseen rooms'' where environments are otherwise new.

\subsection{Inertial Localization Results}

\paragraph{Baselines}
RoNIN \cite{herath2020ronin} learns to predict velocity from IMU signals. We select ResNet-50 as the most effective backbone.
IMUNet \cite{zeinali2024imunet} introduces a new architecture and implements it with various backbones \cite{tan2019mnasnet, tan2019efficientnet}.
NILoc \cite{herath2022neural} directly predicts user location with IMU signals. We choose to use ResNet-18 to encode IMU singals to maintain consistency with our framework.
We train all the aforementioned methods on the EgoExo4D dataset.

\paragraph{Velocity Cumulation versus Direct Location Prediction}
Previous approaches can be divided into two categories.
The first category starts with a given initial localization, estimates velocity using IMU signals, and accumulates the velocity to obtain the target's location, including \cite{herath2020ronin, tan2019mnasnet, tan2019efficientnet, zeinali2024imunet}. The second category directly predicts the target's location, including \cite{herath2022neural} and ours. As shown in \cref{tab:sota_localization}, direct location prediction methods outperform velocity accumulation methods. This is because each recording in the dataset is relatively long, with an average duration of 525 seconds and a maximum of 2,526 seconds. Velocity accumulation methods are significantly affected by error accumulation over such long durations.

\input{./tables/sota_localization}

\begin{figure}[t]
  \centering
   \includegraphics[width=0.9\linewidth]{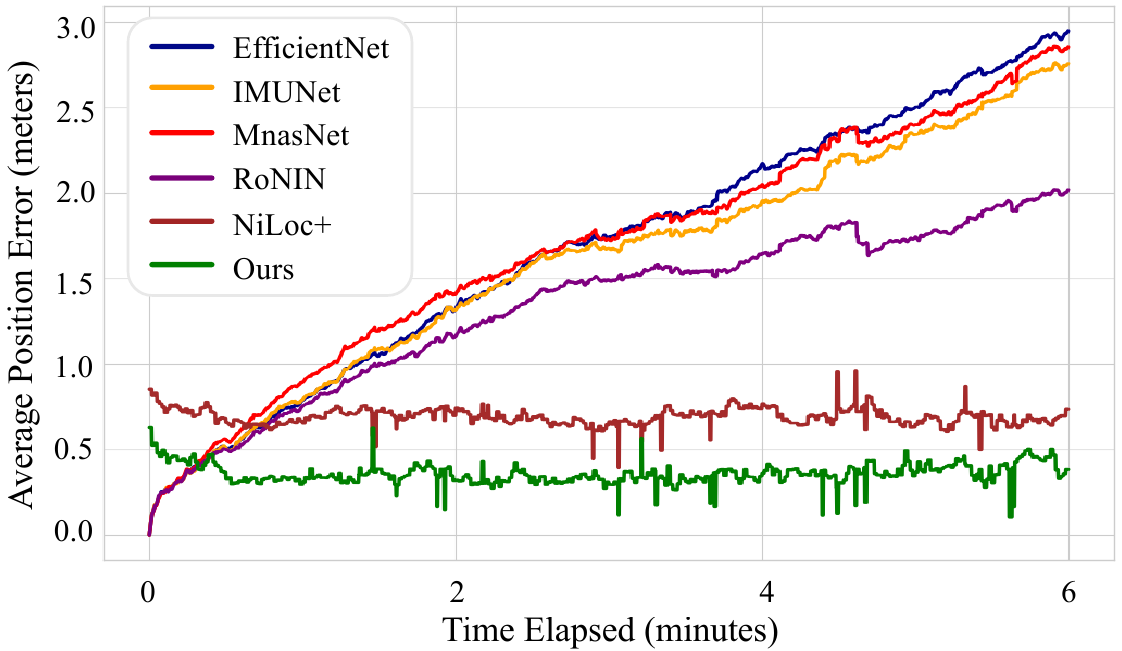}

   \caption{\textbf{Inertial Localization Error Over Time Elapsed}. While velocity accumulation-based methods experience significant trajectory drift, our approach remains accurate over time.}
   \label{fig:sota_localization}
\end{figure}

\paragraph{Trajectory Drift}
\cref{fig:sota_localization} illustrates the difference between the two categories of methods with the change in average position error over elapsed time. In this figure, velocity accumulation methods require a given initial position, resulting in zero error at time zero. However, due to error accumulation over time, their error increases rapidly, surpassing that of our method at around 30 seconds. In contrast, direct localization prediction methods do not require an initial position and maintain a stable error over time.

\paragraph{Direct Localization in Point Clouds}
\looseness=-1
NILoc \cite{herath2022neural} and our method both directly predict a user’s location from IMU signals. However, NILoc relies solely on inertial data, so it can only fit a single scene at a time. In their original work, they trained three separate models, each for a different scene. For a fair comparison, we developed NILoc+, which uses the same multi-scene training data as our Stage~2. Nevertheless, its lack of spatial awareness still leads to reduced accuracy, whereas our approach leverages point cloud structures to deliver robust inertial localization across diverse environments.

\subsection{Inertial Action Recognition Results}
\paragraph{Baselines}
DeepConvLSTM \cite{ordonez2016deep} uses convolutional networks and LSTMs to classify actions from IMU signals.
EVIMAE \cite{zhang2024masked} leverages the MAE framework \cite{he2022masked} to learn representations from both video and multiple IMU signals for action classification. For comparison with our approach, we train a modified version of EVIMAE using data from a single IMU device, employing ViT-Base as the backbone.
IMU2CLIP \cite{moon2023imu2clip} uses a strategy similar to our Stage 1, employing a pretrained CLIP model \cite{radford2021learning, mu2022slip} to guide IMU feature extraction and fine-tuning with an MLP head for action recognition.
For a fair comparison, we use the same backbone for each modality and the same training data as our method when training IMU2CLIP.

\input{./tables/sota_action}

\paragraph{Quantitative Evaluation}
As shown in \cref{tab:sota_action}, DeepConvLSTM performs relatively poorly due to its architecture's difficulty in capturing complex action patterns effectively. Although EVIMAE gains from the advanced reasoning capabilities inherent in transformer networks, it still faces challenges when generalizing to unseen rooms. Much like our method, IMU2CLIP leverages the learning of action semantics from vision-language modalities; however, it doesn't quite match the effectiveness of our approach. This is largely because our method incorporates spatial reasoning, utilizing environmental information to enhance action recognition performance.

\input{./tables/ablation_all}

\begin{figure}[t]
  \centering
   \includegraphics[width=1.\linewidth]{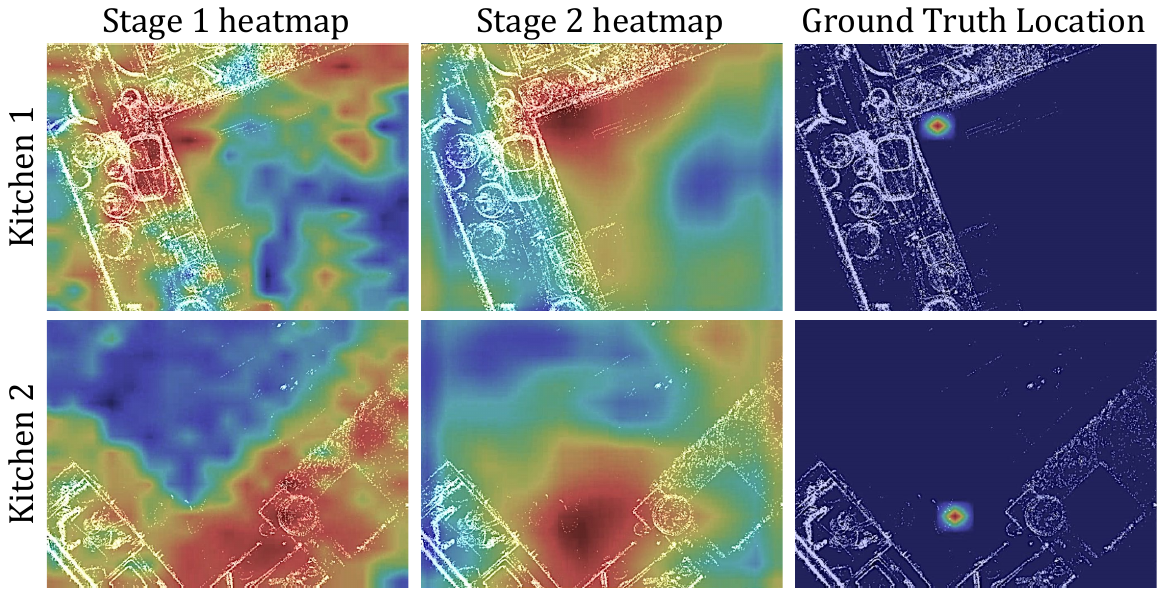}

   \caption{\textbf{Visualization of Heatmaps in Each Stage.}}
   \label{fig:heatmap}
   \vspace{-1em}
\end{figure}

\subsection{Ablation Studies}

\paragraph{Modalities Engagement for Action-aware Alignment}
In Stage~1 of our framework, we focus on effectively training the IMU and the point cloud encoders to yield features that are aligned through action semantics. To accomplish this, we leverage the power of robust pre-trained and pre-aligned vision-language models, such as \cite{radford2021learning, mu2022slip}. These models excel at capturing intricate associations between visual content and corresponding textual descriptions. Our experiments, as detailed in \cref{tab:ablation_main}, affirm the efficacy of incorporating vision and language modalities. By doing so, our model is able to learn and align features across multiple modalities, significantly enhancing both localization and action recognition accuracy. Importantly, it's worth noting that while the integration of vision and language modalities provides substantial benefits during the training phase, they are not required during the inference phase. Furthermore, even in scenarios where action caption annotations are unavailable in the training set, our method does not fail, ensuring reasonable accuracy without relying on complete annotation sets. This demonstrates the broad applicability and flexibility of our approach in real-world environments.

\paragraph{Action Classification Supervision in Stage 2}
We incorporate action classification supervision in Stage 2. The results in \cref{tab:ablation_main} show that including the action loss leads to more accurate location predictions. We believe that explicitly learning action categories helps the model align human motion more effectively with the surrounding environment.

\input{./tables/location-aware-action-rec}

\paragraph{Spatial and Temporal Reasoning in Stage 2}
In the second stage of our framework, we leverage a temporal reasoning module for comprehending a sequence of IMU signals, and a spatial reasoning module for understanding the global environmental point cloud. This dual-layered reasoning framework is essential, as evidenced in \cref{tab:ablation_main}. It shows that when these two modules work in tandem, they align sequential motion data with the environmental settings. This holistic reasoning capability ensures the model generates a more precise and coherent trajectory prediction of the user's movements within their environment.

\vspace{-1em}
\paragraph{Spatial Attention Benefits Inertial Action Recognition}
After obtaining the inertial localization results, we find that incorporating surrounding environmental structure cues is beneficial for inertial action recognition. As shown in \cref{tab:location_aware_action}, using only IMU signals, we achieve results comparable to IMU2CLIP~\cite{moon2023imu2clip}. Naively incorporating global point cloud features with IMU features leads to a performance drop. In contrast, integrating point cloud features with predicted location attention with IMU features provides a clear performance improvement. 

\vspace{-1em}
\paragraph{More Ablation Results in Supplementary Material}
Further ablation results can be found in \cref{tab:ablation_extended} in our Supplementary Material. These include analyses on different vision-language encoders in Stage 1, the preliminary location retrieval accuracy in Stage 1, different architecture designs in Stage 2, and different choices of sequence temporal length in Stage 2.

\subsection{Qualitative evaluations}

\paragraph{Visualization of Heatmaps in Each Stage}
Our framework is capable of producing highly interpretable intermediate results in the form of heatmaps. The heatmap from Stage 1 reflects the direct similarity strength between the features generated by the IMU encoder and the point cloud encoder. The heatmap from Stage 2 represents the localization likelihood at each location, considering both the motion sequence and the global scene structure. As shown in \cref{fig:heatmap}, in Kitchen 1, two peak points appear in the Stage 1 heatmap, indicating that the motion in the IMU signals could plausibly occur at either of these locations. However, after the spatiotemporal reasoning in Stage 2, our framework successfully identifies a single, distinct peak.

\paragraph{Comparison with Previous Methods}

\begin{figure}[t]
  \centering
   \includegraphics[width=\linewidth]{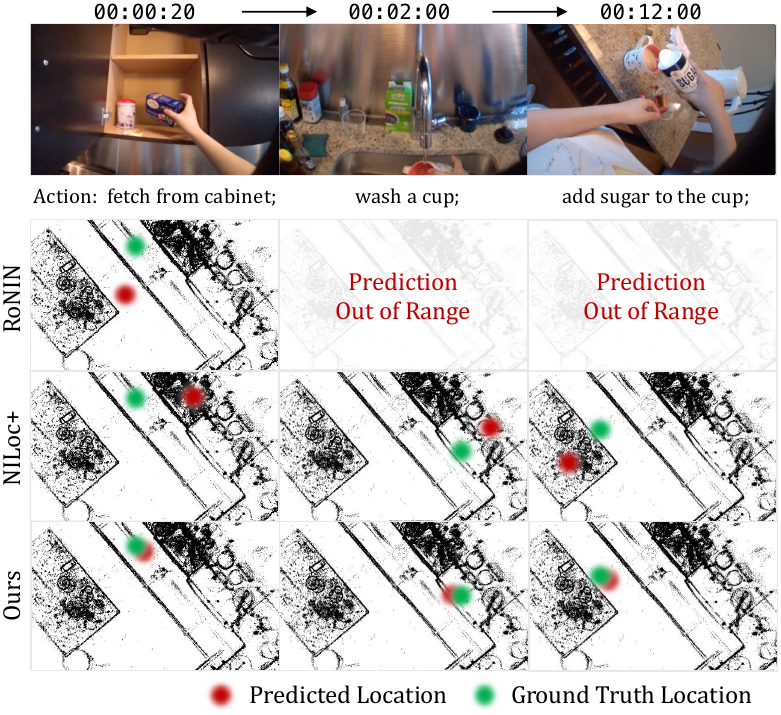}

   \caption{\textbf{Qualitative Comparison with Previous Methods.}}
   \label{fig:result_vis}
\end{figure}

In \cref{fig:result_vis}, we present a qualitative comparison with RoNIN\cite{herath2020ronin} and NILoc\cite{herath2022neural}+. The recorded sequence features a person taking 12 minutes to make a cup of coffee. From the visualization results, we observe that RoNIN suffers from cumulative errors, causing its predicted location to drift outside the point cloud boundary after just 2 minutes. NILoc+, on the other hand, lacks spatial awareness and may produce predictions in physically implausible locations, such as areas already occupied by objects. In contrast, our method leverages action cues such as the upward head motion when fetching items from a cabinet, to establish action-environment correlations, enabling accurate localization.

%% file: tables/sota_localization.tex
\begin{table*}
    \centering
    \vspace{1em}
    \caption{\textbf{Inertial Localization Results}. We evaluate the accuracy using two metrics: the localization success rate (\%) at various error distance thresholds and the Relative Score (RS) metric for localization likelihood prediction (methods that do not generate likelihood predictions are not evaluated with RS). For both metrics, higher values indicate better performance.}
    \label{tab:sota_localization}
    \scalebox{0.8}{
    \begin{tblr}{
      colspec={X[1,l] X[0.1,c] X[1,c] X[1,c] X[1,c] X[1,c] X[0.1,c] X[1,c] X[1,c] X[1,c] X[1,c]},
      cell{1}{1} = {r=2}{l},
      cell{1}{3} = {c=4}{c},
      cell{1}{8} = {c=4}{c},
      hline{1,9} = {-}{0.08em},
      hline{2} = {3-6}{},
      hline{2} = {8-11}{},
      hline{3} = {-}{},
      width=1.1\linewidth,
    }
    Method & & Seen Rooms &       &       &       & & Unseen Rooms &       &       &       \\
           & & 0.2m       & 0.4m  & 0.6m  & RS   & & 0.2m         & 0.4m  & 0.6m  & RS   \\
    MnasNet\cite{tan2019mnasnet}      & & 2.95       & 7.57  & 12.75 & /     & & 2.16         & 5.45  & 9.86  & /     \\
    EfficientNet\cite{tan2019efficientnet} & & 3.10        & 7.77  & 12.71 & /     & & 2.54         & 6.68  & 10.84 & /     \\
    IMUNet\cite{zeinali2024imunet}       & & 3.74       & 10.15 & 17.23 & /     & & 3.40          & 9.17  & 14.63 & /     \\
    RoNIN\cite{herath2020ronin}        & & 4.86       & 12.77 & 20.65 & /     & & 3.96         & 9.52  & 15.65 & /     \\
    NILoc\cite{herath2022neural}$+$     & & 17.03      & 41.31 & 74.15 & 88.17 & & 13.32        & 37.85 & 69.21 & 84.08 \\
    Ours         & &\textbf{ 43.86}      & \textbf{70.15} & \textbf{89.60}  & \textbf{96.01} & & \textbf{26.86}        & \textbf{65.97} & \textbf{90.79} & \textbf{89.55}
    \end{tblr}}
\end{table*}

%% file: tables/sota_action.tex
\begin{table}
    \centering
    \caption{\textbf{Inertial Action Recognition Results}. We evaluate performance using top1 and top5 accuracy metrics. Higher values indicate better performance.}
    \label{tab:sota_action}
    \scalebox{0.8}{
    \begin{tblr}{
      colspec={X[1,l] X[0.1,c] X[1,c] X[1,c] X[0.1,c] X[1,c] X[1,c]}, 
      cell{1}{1} = {r=2}{c}, 
      cell{1}{3} = {c=2}{c}, 
      cell{1}{6} = {c=2}{c}, 
      hline{1,7} = {-}{0.08em},
      hline{2} = {3-4}{}, 
      hline{2} = {6-7}{}, 
      hline{3} = {-}{},
      width=1.25\linewidth, 
    }
    Method       & & Seen Rooms &       & & Unseen Rooms &       \\
                 & & top1       & top5  & & top1         & top5  \\
    {\footnotesize DeepConvLSTM\cite{ordonez2016deep}}     & & 15.20      & 43.27 & & 12.47        & 36.86 \\
    EVIMAE\cite{zhang2024masked}$-$    & & 18.34	    & 46.12	& & 9.30         & 31.10 \\
    IMU2CLIP\cite{moon2023imu2clip}     & & 18.96      & 50.43 & & 12.27        & 37.04 \\
    Ours         & & \textbf{21.48}      & \textbf{53.62} & & \textbf{15.03}        & \textbf{43.34}
    \end{tblr}}
\end{table}

%% file: tables/ablation_all.tex
\begin{table*}
    \centering
    \vspace{1em}
    \caption{\textbf{Ablation Studies}. We report the inertial localization accuracy and the inertial action recognition ($\mathcal{A}$) accuracy simultaneously.}
    \label{tab:ablation_main}
    \scalebox{0.8}{
    \begin{tblr}{
      colspec={X[2.25,l] X[0.1,c] X[1,c] X[1,c] X[1,c] X[1,c] X[1,c] X[1,c] X[0.1,c] X[1,c] X[1,c] X[1,c] X[1,c] X[1,c] X[1,c]},
      cell{1}{1} = {r=2}{l},
      cell{1}{3} = {c=6}{c},
      cell{1}{10} = {c=6}{c},
      cell{3}{1} = {c=15}{l,fg=myblue!50}, 
      cell{6}{1} = {c=15}{l,fg=myblue!50}, 
      cell{9}{1} = {c=15}{l,fg=myblue!50},
      hline{1} = {-}{0.08em},            
      hline{2} = {3-8}{},                
      hline{2} = {10-15}{},              
      hline{3} = {-}{0.08em},            
      hline{6} = {-}{0.08em},            
      hline{9} = {-}{0.08em},           
      hline{13} = {-}{0.08em},           
      width=1.2\linewidth,
    }
    Method & & Seen Rooms &       &       &       &       &       & & Unseen Rooms &       &       &       &       &       \\
           & & 0.2m       & 0.4m  & 0.6m  & RS   & {\small $\mathcal{A}$-top1}  & {\small $\mathcal{A}$-top5}  & & 0.2m         & 0.4m  & 0.6m  & RS   & {\small $\mathcal{A}$-top1}  & {\small $\mathcal{A}$-top5}  \\
           
  \textit{Modalities Engagement (visual and textual action description benefits localization)} & & & & & & & & & & & & & & \\
  {\footnotesize w/o vision language}        & & 39.75 & 66.30 & 87.56 & 95.70 & 13.21 & 51.02 & & 20.41 & 61.95 & 89.83 &  88.13 & 9.53 & 29.09 \\
  {\footnotesize w/ vision language}         & & \textbf{43.86}      & \textbf{70.15} & \textbf{89.60}  & \textbf{96.01} & \textbf{21.48}  & \textbf{53.62}  & & \textbf{26.86}        & \textbf{65.97} & \textbf{90.79} & \textbf{89.55} & \textbf{15.03}  & \textbf{43.34} \\

  \textit{Action Classification Task (explicit action supervision benefits localization)} & & & & & & & & & & & & & & \\
  {\footnotesize w/o action loss}        & & 41.92 & 68.54 & 87.75 & 95.51 & / & / & & 25.37 & 63.79 & 89.28 & 89.02 & / & / \\
  {\footnotesize w/ action loss}       & & \textbf{43.86}      & \textbf{70.15} & \textbf{89.60}  & \textbf{96.01} & \textbf{21.48}  & \textbf{53.62}  & & \textbf{26.86}        & \textbf{65.97} & \textbf{90.79} & \textbf{89.55} & \textbf{15.03}  & \textbf{43.34} \\
  
  \textit{Spatial and Temporal Reasoning} & & & & & & & & & & & & & & \\
  {\footnotesize w/o spatial}  & & 38.68 & 66.13 & 87.78 & 95.05 & 21.20 & 52.96 & & 25.03 & 59.39 & 83.54 &  88.56 & 14.40 & 43.77 \\    
  {\footnotesize w/o temporal}     & & 41.44 & 68.48 & 87.96 & 95.78 & 21.22 & 52.48 & & 26.41 & 64.15 & 89.13 &  89.41 & 14.61 & 43.06 \\
  {\footnotesize w/ both}      & & \textbf{43.86}      & \textbf{70.15} & \textbf{89.60}  & \textbf{96.01} & \textbf{21.48}  & \textbf{53.62}  & & \textbf{26.86}        & \textbf{65.97} & \textbf{90.79} & \textbf{89.55} & \textbf{15.03}  & \textbf{43.34} \\
  
    \end{tblr}}
  \end{table*}

%% file: tables/location-aware-action-rec.tex
\begin{table}
    \centering
    \caption{\textbf{Location-Aware Action Recognition Ablation Study}. ``PC'' denotes point cloud features, and ``LA'' represents location attention.}
    \label{tab:location_aware_action}
    \scalebox{0.8}{
    \begin{tblr}{
      colspec={X[1,l] X[0.1,c] X[1,c] X[1,c] X[0.1,c] X[1,c] X[1,c]}, 
      cell{1}{1} = {r=2}{c}, 
      cell{1}{3} = {c=2}{c}, 
      cell{1}{6} = {c=2}{c}, 
      hline{1,6} = {-}{0.08em},
      hline{2} = {3-4}{}, 
      hline{2} = {6-7}{}, 
      hline{3} = {-}{},
      width=1.25\linewidth, 
    }
    Method       & & Seen Rooms &       & & Unseen Rooms &       \\
                 & & top1       & top5  & & top1         & top5  \\
    w/o PC     & & 18.81      & 50.23 & & 12.37        & 38.21 \\
    w/o LA    & & 17.88	    & 46.22	& & 10.43         & 35.63 \\
    w/ LA PC         & & \textbf{21.48}      & \textbf{53.62} & & \textbf{15.03}        & \textbf{43.34}
    \end{tblr}}
\end{table}

%% file: sec/6_conclusion.tex
\section{Limitations and Future Directions}
While our method can robustly exploit head-mounted IMU signals for human localization within pre-built point clouds, it does hinge on several factors that present avenues for future research. First, the requirement of a complete and up-to-date 3D scan limits applicability in environments subject to frequent layout changes, motivating efforts on incremental or online map updates. Second, our reliance on action-environment correlations makes localization challenging when users remain idle (\eg, standing still for extended periods). Integrating standard inertial navigation solutions and Kalman filters could help bridge gaps during low-motion segments. Finally, our experiments are based on IMU data from head-mounted devices, and substantially different sensor placements (\eg, ankle or wrist) may necessitate model adaptations for robust performance.

\section{Conclusion}
\label{sec:conclusion}

We propose EAIL, a novel framework for inertial localization that uses egocentric action cues from head-mounted IMU signals to improve positioning in 3D point clouds. By aligning short-term IMU data with local structures and incorporating temporal-spatial reasoning, EAIL addresses challenges like trajectory drift and complex actions. Evaluated on the EgoExo4D dataset, it shows strong performance in trajectory prediction and action recognition. This highlights the potential of action-environment correlations as anchors for inertial localization, paving the way for advanced sensing and positioning methods.

\section{Acknowledgement}

This work was supported by 
JSPS KAKENHI Grant Number JP24K02956,
JST ASPIRE Grant Number JPMJAP2303,
JST SPRING Grant Number JPMJSP2108.


%% file: sec/X_suppl.tex
\clearpage
\setcounter{page}{1}
\maketitlesupplementary

\input{./tables/ablation_supp}

\section{Supplementary Ablation Studies}
\label{sec:SAS}

\paragraph{Different Vision-Language Encoder in Stage 1}  
We compare CLIP-Large\cite{radford2021learning}, InternVL\cite{chen2024internvl}, and SLIP-Base\cite{mu2022slip} as possible backbones for the vision-language encoders. While InternVL scores the highest on some metrics, SLIP generally attains stronger localization accuracy. Hence, in our default configuration, we adopt SLIP as the vision-language encoder.

\paragraph{Temporal Length of Input IMU Sequence in Stage 2}  
We experiment with sequences of 5s, 10s, and 20s of IMU data. Shorter 5s windows excel on the seen rooms but underperform on the unseen set. Longer 20s windows better preserve temporal context and yield a slight gain in unseen action top‐1 performance, yet they degrade localization in both settings. By contrast, 10s serves as a balanced choice, delivering robust localization and action recognition across seen/unseen scenarios.

\paragraph{Residual Connection in Stage 2 Architecture Design}  
Removing residual connections, either on the IMU feature path or on the point cloud feature path, consistently degrades results. In contrast, preserving both forms of residual connections significantly improves localization and action accuracy. This highlights the importance of allowing the network to fuse new spatiotemporal cues while maintaining a direct pathway for unimodal features.

\paragraph{Preliminary Location Retrieval Accuracy in Stage 1}  
Lastly, we assess how well the short-term feature alignment from Stage 1 alone can recover the user’s location. Doing so attains considerably lower accuracy compared to our full Stage 2 inference. This underscores the utility of leveraging both temporal context and global spatial reasoning to refine the initial retrieval from Stage 1.

\section{Supplementary Video}

We create a supplementary video that further illustrates the performance of our framework. The heatmaps featured in the video are interpreted similarly to those shown in \cref{fig:heatmap}, providing visual insight into the localization process. The video demonstrates the robustness of our approach, highlighting its capability for accurate long-term human tracking across extended periods.

%% file: tables/ablation_supp.tex
\begin{table*}[b]
    \centering
    \caption{Supplementary Ablation Studies.}
    \label{tab:ablation_extended}
    \scalebox{0.9}{
    \begin{tblr}{
      colspec={X[2.25,l] X[0.1,c] X[1,c] X[1,c] X[1,c] X[1,c] X[1,c] X[1,c] X[0.1,c] X[1,c] X[1,c] X[1,c] X[1,c] X[1,c] X[1,c]},
      cell{1}{1} = {r=2}{l},
      cell{1}{3} = {c=6}{c},
      cell{1}{10} = {c=6}{c},
      cell{3}{1} = {c=15}{l,fg=myblue!50},
      cell{7}{1} = {c=15}{l,fg=myblue!50},
      cell{11}{1} = {c=15}{l,fg=myblue!50},
      cell{15}{1} = {c=15}{l,fg=myblue!50},
      hline{1} = {-}{0.08em},
      hline{2} = {3-8}{},
      hline{2} = {10-15}{},
      hline{3} = {-}{0.08em},
      hline{7} = {-}{0.08em},
      hline{11} = {-}{0.08em},
      hline{15} = {-}{0.08em},
      hline{18} = {-}{0.08em},
      width=1.1\linewidth,
    }
    Method & & Seen Rooms &       &       &       &       &       & & Unseen Rooms &       &       &       &       &       \\
           & & 0.2m       & 0.4m  & 0.6m  & RS   & {\footnotesize $\mathcal{A}$-top1}  & {\footnotesize $\mathcal{A}$-top5}  & & 0.2m         & 0.4m  & 0.6m  & RS   & {\footnotesize $\mathcal{A}$-top1}  & {\footnotesize $\mathcal{A}$-top5}  \\

\textit{Different Vision-Language Encoder in Stage 1} & & & & & & & & & & & & & & \\
{\footnotesize CLIP}      & & 42.72 & 69.33 & 89.63 & 95.94 & 20.81 & 52.05 & & 21.60 & 53.45 & 83.41 & 86.44 & 12.68 & 40.38 \\
{\footnotesize InternVL}  & & 42.41 & 69.21 & \textbf{90.95} & \textbf{96.15} & 17.15 & 50.80 & & 26.11 & 60.46 & 87.70 & \textbf{90.18} & 9.84 & 39.44 \\
{\footnotesize SLIP}      & & \textbf{43.86} & \textbf{70.15} & 89.6 & 96.01 & \textbf{21.48} & \textbf{53.62} & & \textbf{26.86} & \textbf{65.97} & \textbf{90.79} & 89.55 & \textbf{15.03} & \textbf{43.34} \\

\textit{Temporal Length of Input IMU Sequence in Stage 2} & & & & & & & & & & & & & & \\
{\footnotesize 5s}      & & \textbf{45.27} & \textbf{71.86} & \textbf{90.84} & \textbf{96.18} & 21.23 & 53.57 & & 25.69 & 64.07 & 87.11 & 89.28 & 14.53 & 43.62 \\
{\footnotesize 20s}     & & 43.10 & 69.61 & 89.11 & 95.84 & 20.82 & 52.53 & & 24.77 & 61.09 & 85.99 & 89.15 & \textbf{15.41} & \textbf{44.77} \\
{\footnotesize 10s}     & & 43.86 & 70.15 & 89.6 & 96.01 & \textbf{21.48} & \textbf{53.62} & & \textbf{26.86} & \textbf{65.97} & \textbf{90.79} & \textbf{89.55} & 15.03 & 43.34 \\

\textit{Residual Connection in Stage 2 Architecture Design} & & & & & & & & & & & & & & \\
{\footnotesize w/o IMU residual}  & & 36.29 & 63.56 & 86.84 & 95.05 & 21.45 & 52.38 & & 26.07 & 60.41 & 86.03 & 89.10 & 14.90 & 43.15 \\
{\footnotesize w/o PC residual}   & & 39.14 & 64.71 & 86.16 & 95.30 & 20.24 & 52.08 & & 23.51 & 59.66 & 86.02 & 87.78 & 14.60 & 43.77 \\
{\footnotesize w/ both}           & & \textbf{43.86} & \textbf{70.15} & \textbf{89.6} & \textbf{96.01} & \textbf{21.48} & \textbf{53.62} & & \textbf{26.86} & \textbf{65.97} & \textbf{90.79} & \textbf{89.55} & \textbf{15.03} & \textbf{43.34} \\

\textit{Preliminary Location Retrieval Accuracy in Stage 1} & & & & & & & & & & & & & & \\
{\footnotesize Stage 1 Retrieval} & & 18.29 & 41.23 & 66.70 & 85.47 & / & / & & 14.25 & 38.51 & 64.31 & 78.31 & / & / \\
{\footnotesize Full 2 Stages}           & & \textbf{43.86} & \textbf{70.15} & \textbf{89.6} & \textbf{96.01} & \textbf{21.48} & \textbf{53.62} & & \textbf{26.86} & \textbf{65.97} & \textbf{90.79} & \textbf{89.55} & \textbf{15.03} & \textbf{43.34} \\
    \end{tblr}}
\end{table*}

%% file: main.bbl
\begin{thebibliography}{70}
\providecommand{\natexlab}[1]{#1}
\providecommand{\url}[1]{\texttt{#1}}
\expandafter\ifx\csname urlstyle\endcsname\relax
  \providecommand{\doi}[1]{doi: #1}\else
  \providecommand{\doi}{doi: \begingroup \urlstyle{rm}\Url}\fi

\bibitem[sca()]{scaniverse}
Scaniverse.
\newblock Accessed on March 7th, 2025.

\bibitem[Ashutosh et~al.(2023)Ashutosh, Girdhar, Torresani, and Grauman]{Ashutosh_2023_CVPR}
Kumar Ashutosh, Rohit Girdhar, Lorenzo Torresani, and Kristen Grauman.
\newblock Hiervl: Learning hierarchical video-language embeddings.
\newblock In \emph{Proceedings of the IEEE/CVF Conference on Computer Vision and Pattern Recognition (CVPR)}, pages 23066--23078, 2023.

\bibitem[Brajdic and Harle(2013)]{brajdic2013walk}
Agata Brajdic and Robert Harle.
\newblock Walk detection and step counting on unconstrained smartphones.
\newblock In \emph{Proceedings of the 2013 ACM international joint conference on Pervasive and ubiquitous computing}, pages 225--234, 2013.

\bibitem[Chen et~al.(2018{\natexlab{a}})Chen, Lu, Markham, and Trigoni]{chen2018ionet}
Changhao Chen, Xiaoxuan Lu, Andrew Markham, and Niki Trigoni.
\newblock Ionet: Learning to cure the curse of drift in inertial odometry.
\newblock In \emph{Proceedings of the AAAI Conference on Artificial Intelligence}, 2018{\natexlab{a}}.

\bibitem[Chen et~al.(2018{\natexlab{b}})Chen, Zhao, Lu, Wang, Markham, and Trigoni]{chen2018oxiod}
Changhao Chen, Peijun Zhao, Chris~Xiaoxuan Lu, Wei Wang, Andrew Markham, and Niki Trigoni.
\newblock Oxiod: The dataset for deep inertial odometry.
\newblock \emph{arXiv preprint arXiv:1809.07491}, 2018{\natexlab{b}}.

\bibitem[Chen et~al.(2024{\natexlab{a}})Chen, Ashutosh, Girdhar, Harwath, and Grauman]{chen2024soundingactions}
Changan Chen, Kumar Ashutosh, Rohit Girdhar, David Harwath, and Kristen Grauman.
\newblock Soundingactions: Learning how actions sound from narrated egocentric videos.
\newblock In \emph{Proceedings of the IEEE/CVF Conference on Computer Vision and Pattern Recognition (CVPR)}, pages 27252--27262, 2024{\natexlab{a}}.

\bibitem[Chen et~al.(2024{\natexlab{b}})Chen, Wu, Wang, Su, Chen, Xing, Zhong, Zhang, Zhu, Lu, et~al.]{chen2024internvl}
Zhe Chen, Jiannan Wu, Wenhai Wang, Weijie Su, Guo Chen, Sen Xing, Muyan Zhong, Qinglong Zhang, Xizhou Zhu, Lewei Lu, et~al.
\newblock Internvl: Scaling up vision foundation models and aligning for generic visual-linguistic tasks.
\newblock In \emph{Proceedings of the IEEE/CVF Conference on Computer Vision and Pattern Recognition (CVPR)}, pages 24185--24198, 2024{\natexlab{b}}.

\bibitem[Damen et~al.(2018)Damen, Doughty, Farinella, Fidler, Furnari, Kazakos, Moltisanti, Munro, Perrett, Price, et~al.]{damen2018scaling}
Dima Damen, Hazel Doughty, Giovanni~Maria Farinella, Sanja Fidler, Antonino Furnari, Evangelos Kazakos, Davide Moltisanti, Jonathan Munro, Toby Perrett, Will Price, et~al.
\newblock Scaling egocentric vision: The epic-kitchens dataset.
\newblock In \emph{Proceedings of the European Conference on Computer Vision (ECCV)}, pages 720--736, 2018.

\bibitem[Delitzas et~al.(2024)Delitzas, Takmaz, Tombari, Sumner, Pollefeys, and Engelmann]{delitzas2024scenefun3d}
Alexandros Delitzas, Ayca Takmaz, Federico Tombari, Robert Sumner, Marc Pollefeys, and Francis Engelmann.
\newblock Scenefun3d: fine-grained functionality and affordance understanding in 3d scenes.
\newblock In \emph{Proceedings of the IEEE/CVF Conference on Computer Vision and Pattern Recognition (CVPR)}, pages 14531--14542, 2024.

\bibitem[Dosovitskiy et~al.(2020)Dosovitskiy, Beyer, Kolesnikov, Weissenborn, Zhai, Unterthiner, Dehghani, Minderer, Heigold, Gelly, et~al.]{dosovitskiy2020image}
Alexey Dosovitskiy, Lucas Beyer, Alexander Kolesnikov, Dirk Weissenborn, Xiaohua Zhai, Thomas Unterthiner, Mostafa Dehghani, Matthias Minderer, Georg Heigold, Sylvain Gelly, et~al.
\newblock An image is worth 16x16 words: Transformers for image recognition at scale.
\newblock \emph{arXiv preprint arXiv:2010.11929}, 2020.

\bibitem[Dou et~al.(2024)Dou, Yang, Nagarajan, Wang, Huang, Peng, Kitani, and Chu]{dou2024unlocking}
Zi-Yi Dou, Xitong Yang, Tushar Nagarajan, Huiyu Wang, Jing Huang, Nanyun Peng, Kris Kitani, and Fu-Jen Chu.
\newblock Unlocking exocentric video-language data for egocentric video representation learning.
\newblock \emph{arXiv preprint arXiv:2408.03567}, 2024.

\bibitem[Engel et~al.(2023)Engel, Somasundaram, Goesele, Sun, Gamino, Turner, Talattof, Yuan, Souti, Meredith, et~al.]{engel2023project}
Jakob Engel, Kiran Somasundaram, Michael Goesele, Albert Sun, Alexander Gamino, Andrew Turner, Arjang Talattof, Arnie Yuan, Bilal Souti, Brighid Meredith, et~al.
\newblock Project aria: A new tool for egocentric multi-modal ai research.
\newblock \emph{arXiv preprint arXiv:2308.13561}, 2023.

\bibitem[et~al.(2012)]{hardegger2012actionslam}
Hardegger et al.
\newblock Actionslam: Using location-related actions as landmarks in pedestrian slam.
\newblock In \emph{IPIN}, 2012.

\bibitem[et~al.(2022)]{zhang2022positioning}
Zhang et al.
\newblock Positioning method of pedestrian dead reckoning based on human activity recognition.
\newblock In \emph{IPIN}, 2022.

\bibitem[Fan(2019)]{fan2019egovqa}
Chenyou Fan.
\newblock Egovqa-an egocentric video question answering benchmark dataset.
\newblock In \emph{Proceedings of the IEEE/CVF International Conference on Computer Vision Workshops}, pages 0--0, 2019.

\bibitem[Gong et~al.(2023)Gong, Mohan, Dhingra, Bazin, Li, Wang, and Ranjan]{gong2023mmg}
Xinyu Gong, Sreyas Mohan, Naina Dhingra, Jean-Charles Bazin, Yilei Li, Zhangyang Wang, and Rakesh Ranjan.
\newblock Mmg-ego4d: Multimodal generalization in egocentric action recognition.
\newblock In \emph{Proceedings of the IEEE/CVF Conference on Computer Vision and Pattern Recognition (CVPR)}, pages 6481--6491, 2023.

\bibitem[Grauman et~al.(2022)Grauman, Westbury, Byrne, Chavis, Furnari, Girdhar, Hamburger, Jiang, Liu, Liu, et~al.]{grauman2022ego4d}
Kristen Grauman, Andrew Westbury, Eugene Byrne, Zachary Chavis, Antonino Furnari, Rohit Girdhar, Jackson Hamburger, Hao Jiang, Miao Liu, Xingyu Liu, et~al.
\newblock Ego4d: Around the world in 3,000 hours of egocentric video.
\newblock In \emph{Proceedings of the IEEE/CVF Conference on Computer Vision and Pattern Recognition (CVPR)}, pages 18995--19012, 2022.

\bibitem[Grauman et~al.(2024)Grauman, Westbury, Torresani, Kitani, Malik, Afouras, Ashutosh, Baiyya, Bansal, Boote, et~al.]{grauman2024ego}
Kristen Grauman, Andrew Westbury, Lorenzo Torresani, Kris Kitani, Jitendra Malik, Triantafyllos Afouras, Kumar Ashutosh, Vijay Baiyya, Siddhant Bansal, Bikram Boote, et~al.
\newblock Ego-exo4d: Understanding skilled human activity from first-and third-person perspectives.
\newblock In \emph{Proceedings of the IEEE/CVF Conference on Computer Vision and Pattern Recognition (CVPR)}, pages 19383--19400, 2024.

\bibitem[Gu et~al.(2024)Gu, Kuwajerwala, Morin, Jatavallabhula, Sen, Agarwal, Rivera, Paul, Ellis, Chellappa, et~al.]{gu2024conceptgraphs}
Qiao Gu, Ali Kuwajerwala, Sacha Morin, Krishna~Murthy Jatavallabhula, Bipasha Sen, Aditya Agarwal, Corban Rivera, William Paul, Kirsty Ellis, Rama Chellappa, et~al.
\newblock Conceptgraphs: Open-vocabulary 3d scene graphs for perception and planning.
\newblock In \emph{IEEE International Conference on Robotics and Automation (ICRA)}, pages 5021--5028. IEEE, 2024.

\bibitem[He et~al.(2016)He, Zhang, Ren, and Sun]{he2016deep}
Kaiming He, Xiangyu Zhang, Shaoqing Ren, and Jian Sun.
\newblock Deep residual learning for image recognition.
\newblock In \emph{Proceedings of the IEEE/CVF Conference on Computer Vision and Pattern Recognition (CVPR)}, pages 770--778, 2016.

\bibitem[He et~al.(2022)He, Chen, Xie, Li, Doll{\'a}r, and Girshick]{he2022masked}
Kaiming He, Xinlei Chen, Saining Xie, Yanghao Li, Piotr Doll{\'a}r, and Ross Girshick.
\newblock Masked autoencoders are scalable vision learners.
\newblock In \emph{Proceedings of the IEEE/CVF Conference on Computer Vision and Pattern Recognition (CVPR)}, pages 16000--16009, 2022.

\bibitem[Herath et~al.(2020)Herath, Yan, and Furukawa]{herath2020ronin}
Sachini Herath, Hang Yan, and Yasutaka Furukawa.
\newblock Ronin: Robust neural inertial navigation in the wild: Benchmark, evaluations, \& new methods.
\newblock In \emph{IEEE International Conference on Robotics and Automation (ICRA)}, pages 3146--3152. IEEE, 2020.

\bibitem[Herath et~al.(2021)Herath, Irandoust, Chen, Qian, Kim, and Furukawa]{herath2021fusion}
Sachini Herath, Saghar Irandoust, Bowen Chen, Yiming Qian, Pyojin Kim, and Yasutaka Furukawa.
\newblock Fusion-dhl: Wifi, imu, and floorplan fusion for dense history of locations in indoor environments.
\newblock In \emph{IEEE International Conference on Robotics and Automation (ICRA)}, pages 5677--5683. IEEE, 2021.

\bibitem[Herath et~al.(2022)Herath, Caruso, Liu, Chen, and Furukawa]{herath2022neural}
Sachini Herath, David Caruso, Chen Liu, Yufan Chen, and Yasutaka Furukawa.
\newblock Neural inertial localization.
\newblock In \emph{Proceedings of the IEEE/CVF Conference on Computer Vision and Pattern Recognition (CVPR)}, pages 6604--6613, 2022.

\bibitem[Huang et~al.(2024{\natexlab{a}})Huang, Chen, Xu, Zhang, Yang, Pei, Zhang, Dong, Wang, Wang, et~al.]{huang2024egoexolearn}
Yifei Huang, Guo Chen, Jilan Xu, Mingfang Zhang, Lijin Yang, Baoqi Pei, Hongjie Zhang, Lu Dong, Yali Wang, Limin Wang, et~al.
\newblock Egoexolearn: A dataset for bridging asynchronous ego-and exo-centric view of procedural activities in real world.
\newblock In \emph{Proceedings of the IEEE/CVF Conference on Computer Vision and Pattern Recognition (CVPR)}, pages 22072--22086, 2024{\natexlab{a}}.

\bibitem[Huang et~al.(2024{\natexlab{b}})Huang, Xu, Pei, He, Chen, Yang, Chen, Wang, Nie, Liu, et~al.]{huang2024vinci}
Yifei Huang, Jilan Xu, Baoqi Pei, Yuping He, Guo Chen, Lijin Yang, Xinyuan Chen, Yaohui Wang, Zheng Nie, Jinyao Liu, et~al.
\newblock Vinci: A real-time embodied smart assistant based on egocentric vision-language model.
\newblock \emph{arXiv preprint arXiv:2412.21080}, 2024{\natexlab{b}}.

\bibitem[Islam et~al.(2024)Islam, Ho, Yang, Nagarajan, Torresani, and Bertasius]{islam2024video}
Md~Mohaiminul Islam, Ngan Ho, Xitong Yang, Tushar Nagarajan, Lorenzo Torresani, and Gedas Bertasius.
\newblock Video recap: Recursive captioning of hour-long videos.
\newblock In \emph{Proceedings of the IEEE/CVF Conference on Computer Vision and Pattern Recognition (CVPR)}, pages 18198--18208, 2024.

\bibitem[Kendall et~al.(2015)Kendall, Grimes, and Cipolla]{kendall2015posenet}
Alex Kendall, Matthew Grimes, and Roberto Cipolla.
\newblock Posenet: A convolutional network for real-time 6-dof camera relocalization.
\newblock In \emph{Proceedings of the IEEE International Conference on Computer Vision (ICCV)}, pages 2938--2946, 2015.

\bibitem[Kim and Shin(2021)]{kim2021deep}
Kwan-Soo Kim and Yoan Shin.
\newblock Deep learning-based pdr scheme that fuses smartphone sensors and gps location changes.
\newblock \emph{IEEE Access}, 9:\penalty0 158616--158631, 2021.

\bibitem[Labb{\'e} and Michaud(2019)]{labbe2019rtab}
Mathieu Labb{\'e} and Fran{\c{c}}ois Michaud.
\newblock Rtab-map as an open-source lidar and visual simultaneous localization and mapping library for large-scale and long-term online operation.
\newblock \emph{Journal of Field Robotics}, 36\penalty0 (2):\penalty0 416--446, 2019.

\bibitem[Leutenegger et~al.(2015)Leutenegger, Lynen, Bosse, Siegwart, and Furgale]{leutenegger2015keyframe}
Stefan Leutenegger, Simon Lynen, Michael Bosse, Roland Siegwart, and Paul Furgale.
\newblock Keyframe-based visual--inertial odometry using nonlinear optimization.
\newblock \emph{The International Journal of Robotics Research}, 34\penalty0 (3):\penalty0 314--334, 2015.

\bibitem[Li et~al.(2015)Li, Wang, and Liu]{li2015bluetooth}
Xin Li, Jian Wang, and Chunyan Liu.
\newblock A bluetooth/pdr integration algorithm for an indoor positioning system.
\newblock \emph{Sensors}, 15\penalty0 (10):\penalty0 24862--24885, 2015.

\bibitem[Lin et~al.(2022)Lin, Wang, Soldan, Wray, Yan, Xu, Gao, Tu, Zhao, Kong, et~al.]{kevin2022egovlp}
Kevin~Qinghong Lin, Alex~Jinpeng Wang, Mattia Soldan, Michael Wray, Rui Yan, Eric~Zhongcong Xu, Difei Gao, Rongcheng Tu, Wenzhe Zhao, Weijie Kong, et~al.
\newblock Egocentric video-language pretraining.
\newblock \emph{arXiv preprint arXiv:2206.01670}, 2022.

\bibitem[Lin et~al.(2025)Lin, Ohkawa, Huang, Zhang, Cai, Li, Furuta, and Sato]{lin2025simhand}
Nie Lin, Takehiko Ohkawa, Yifei Huang, Mingfang Zhang, Minjie Cai, Ming Li, Ryosuke Furuta, and Yoichi Sato.
\newblock Simhand: Mining similar hands for large-scale 3d hand pose pre-training.
\newblock \emph{arXiv preprint arXiv:2502.15251}, 2025.

\bibitem[Liu et~al.(2024{\natexlab{a}})Liu, Yin, Kaufmann, He, Christen, Song, and Hui]{liu2024egohdm}
Bonan Liu, Handi Yin, Manuel Kaufmann, Jinhao He, Sammy Christen, Jie Song, and Pan Hui.
\newblock Egohdm: An online egocentric-inertial human motion capture, localization, and dense mapping system.
\newblock \emph{arXiv preprint arXiv:2409.00343}, 2024{\natexlab{a}}.

\bibitem[Liu et~al.(2024{\natexlab{b}})Liu, Ohkawa, Zhang, and Sato]{liu2024single}
Ruicong Liu, Takehiko Ohkawa, Mingfang Zhang, and Yoichi Sato.
\newblock Single-to-dual-view adaptation for egocentric 3d hand pose estimation.
\newblock In \emph{Proceedings of the IEEE/CVF Conference on Computer Vision and Pattern Recognition}, pages 677--686, 2024{\natexlab{b}}.

\bibitem[Liu et~al.(2020)Liu, Caruso, Ilg, Dong, Mourikis, Daniilidis, Kumar, and Engel]{liu2020tlio}
Wenxin Liu, David Caruso, Eddy Ilg, Jing Dong, Anastasios~I Mourikis, Kostas Daniilidis, Vijay Kumar, and Jakob Engel.
\newblock Tlio: Tight learned inertial odometry.
\newblock \emph{IEEE Robotics and Automation Letters}, 5\penalty0 (4):\penalty0 5653--5660, 2020.

\bibitem[Liu et~al.(2024{\natexlab{c}})Liu, Sakuma, and Okutomi]{liu2024vsrd}
Zihua Liu, Hiroki Sakuma, and Masatoshi Okutomi.
\newblock Vsrd: Instance-aware volumetric silhouette rendering for weakly supervised 3d object detection.
\newblock In \emph{Proceedings of the IEEE/CVF Conference on Computer Vision and Pattern Recognition}, pages 17354--17363, 2024{\natexlab{c}}.

\bibitem[Lowe(2004)]{lowe2004distinctive}
David~G Lowe.
\newblock Distinctive image features from scale-invariant keypoints.
\newblock \emph{International Journal of Computer Vision}, 60:\penalty0 91--110, 2004.

\bibitem[Lu et~al.(2023)Lu, Chang, Jing, Boularias, and Bekris]{lu2023ovir}
Shiyang Lu, Haonan Chang, Eric~Pu Jing, Abdeslam Boularias, and Kostas Bekris.
\newblock Ovir-3d: Open-vocabulary 3d instance retrieval without training on 3d data.
\newblock In \emph{Conference on Robot Learning}, pages 1610--1620. PMLR, 2023.

\bibitem[Moon et~al.(2023)Moon, Madotto, Lin, Saraf, Bearman, and Damavandi]{moon2023imu2clip}
Seungwhan Moon, Andrea Madotto, Zhaojiang Lin, Aparajita Saraf, Amy Bearman, and Babak Damavandi.
\newblock Imu2clip: language-grounded motion sensor translation with multimodal contrastive learning.
\newblock In \emph{Findings of the Association for Computational Linguistics: EMNLP 2023}, pages 13246--13253, 2023.

\bibitem[Mourikis and Roumeliotis(2007)]{mourikis2007multi}
Anastasios~I Mourikis and Stergios~I Roumeliotis.
\newblock A multi-state constraint kalman filter for vision-aided inertial navigation.
\newblock In \emph{IEEE International Conference on Robotics and Automation (ICRA)}, pages 3565--3572. IEEE, 2007.

\bibitem[Mu et~al.(2022)Mu, Kirillov, Wagner, and Xie]{mu2022slip}
Norman Mu, Alexander Kirillov, David Wagner, and Saining Xie.
\newblock Slip: Self-supervision meets language-image pre-training.
\newblock In \emph{Proceedings of the European Conference on Computer Vision (ECCV)}, pages 529--544. Springer, 2022.

\bibitem[Ord{\'o}{\~n}ez and Roggen(2016)]{ordonez2016deep}
Francisco~Javier Ord{\'o}{\~n}ez and Daniel Roggen.
\newblock Deep convolutional and lstm recurrent neural networks for multimodal wearable activity recognition.
\newblock \emph{Sensors}, 16\penalty0 (1):\penalty0 115, 2016.

\bibitem[Park et~al.(2021)Park, Bokijonov, and Choi]{park2021review}
Sebeom Park, Shokhrukh Bokijonov, and Yosoon Choi.
\newblock Review of microsoft hololens applications over the past five years.
\newblock \emph{Applied sciences}, 11\penalty0 (16):\penalty0 7259, 2021.

\bibitem[Pei et~al.(2025)Pei, Huang, Xu, Chen, He, Yang, Wang, Xie, Qiao, Wu, and Wang]{pei2025modeling}
Baoqi Pei, Yifei Huang, Jilan Xu, Guo Chen, Yuping He, Lijin Yang, Yali Wang, Weidi Xie, Yu Qiao, Fei Wu, and Limin Wang.
\newblock Modeling fine-grained hand-object dynamics for egocentric video representation learning.
\newblock In \emph{International Conference on Learning Representations}, 2025.

\bibitem[Pramanick et~al.(2023)Pramanick, Song, Nag, Lin, Shah, Shou, Chellappa, and Zhang]{pramanick2023egovlpv2}
Shraman Pramanick, Yale Song, Sayan Nag, Kevin~Qinghong Lin, Hardik Shah, Mike~Zheng Shou, Rama Chellappa, and Pengchuan Zhang.
\newblock Egovlpv2: Egocentric video-language pre-training with fusion in the backbone.
\newblock In \emph{Proceedings of the IEEE/CVF Conference on Computer Vision and Pattern Recognition (CVPR)}, pages 5285--5297, 2023.

\bibitem[Qi et~al.(2017)Qi, Yi, Su, and Guibas]{qi2017pointnet++}
Charles~Ruizhongtai Qi, Li Yi, Hao Su, and Leonidas~J Guibas.
\newblock Pointnet++: Deep hierarchical feature learning on point sets in a metric space.
\newblock \emph{Advances in Neural Information Processing Systems}, 30, 2017.

\bibitem[Qin et~al.(2018)Qin, Li, and Shen]{qin2018vins}
Tong Qin, Peiliang Li, and Shaojie Shen.
\newblock Vins-mono: A robust and versatile monocular visual-inertial state estimator.
\newblock \emph{IEEE Transactions on Robotics}, 34\penalty0 (4):\penalty0 1004--1020, 2018.

\bibitem[Radford et~al.(2021)Radford, Kim, Hallacy, Ramesh, Goh, Agarwal, Sastry, Askell, Mishkin, Clark, et~al.]{radford2021learning}
Alec Radford, Jong~Wook Kim, Chris Hallacy, Aditya Ramesh, Gabriel Goh, Sandhini Agarwal, Girish Sastry, Amanda Askell, Pamela Mishkin, Jack Clark, et~al.
\newblock Learning transferable visual models from natural language supervision.
\newblock In \emph{International Conference on Machine Learning}, pages 8748--8763. PmLR, 2021.

\bibitem[Sarlin et~al.(2022)Sarlin, Dusmanu, Sch{\"o}nberger, Speciale, Gruber, Larsson, Miksik, and Pollefeys]{sarlin2022lamar}
Paul-Edouard Sarlin, Mihai Dusmanu, Johannes~L Sch{\"o}nberger, Pablo Speciale, Lukas Gruber, Viktor Larsson, Ondrej Miksik, and Marc Pollefeys.
\newblock Lamar: Benchmarking localization and mapping for augmented reality.
\newblock In \emph{Proceedings of the European Conference on Computer Vision (ECCV)}, pages 686--704. Springer, 2022.

\bibitem[Shen et~al.(2024)Shen, Wang, Yang, Feiszli, Elhamifar, Torresani, and Mavroudi]{shen2024learning}
Yuhan Shen, Huiyu Wang, Xitong Yang, Matt Feiszli, Ehsan Elhamifar, Lorenzo Torresani, and Effrosyni Mavroudi.
\newblock Learning to segment referred objects from narrated egocentric videos.
\newblock In \emph{Proceedings of the IEEE/CVF Conference on Computer Vision and Pattern Recognition (CVPR)}, pages 14510--14520, 2024.

\bibitem[Shu et~al.(2015)Shu, Shin, He, and Chen]{shu2015last}
Yuanchao Shu, Kang~G Shin, Tian He, and Jiming Chen.
\newblock Last-mile navigation using smartphones.
\newblock In \emph{Proceedings of the 21st annual international conference on mobile computing and networking}, pages 512--524, 2015.

\bibitem[Sun et~al.(2021)Sun, Melamed, and Kitani]{sun2021idol}
Scott Sun, Dennis Melamed, and Kris Kitani.
\newblock Idol: Inertial deep orientation-estimation and localization.
\newblock In \emph{Proceedings of the AAAI Conference on Artificial Intelligence}, pages 6128--6137, 2021.

\bibitem[Takmaz et~al.(2023)Takmaz, Fedele, Sumner, Pollefeys, Tombari, and Engelmann]{takmaz2023openmask3d}
Ay{\c{c}}a Takmaz, Elisabetta Fedele, Robert~W Sumner, Marc Pollefeys, Federico Tombari, and Francis Engelmann.
\newblock Openmask3d: Open-vocabulary 3d instance segmentation.
\newblock \emph{arXiv preprint arXiv:2306.13631}, 2023.

\bibitem[Tan and Le(2019)]{tan2019efficientnet}
Mingxing Tan and Quoc Le.
\newblock Efficientnet: Rethinking model scaling for convolutional neural networks.
\newblock In \emph{International Conference on Machine Learning}, pages 6105--6114. PMLR, 2019.

\bibitem[Tan et~al.(2019)Tan, Chen, Pang, Vasudevan, Sandler, Howard, and Le]{tan2019mnasnet}
Mingxing Tan, Bo Chen, Ruoming Pang, Vijay Vasudevan, Mark Sandler, Andrew Howard, and Quoc~V Le.
\newblock Mnasnet: Platform-aware neural architecture search for mobile.
\newblock In \emph{Proceedings of the IEEE/CVF Conference on Computer Vision and Pattern Recognition (CVPR)}, pages 2820--2828, 2019.

\bibitem[Titterton and Weston(2004)]{titterton2004strapdown}
David Titterton and John~L Weston.
\newblock \emph{Strapdown inertial navigation technology}.
\newblock IET, 2004.

\bibitem[Vaswani et~al.(2017)Vaswani, Shazeer, Parmar, Uszkoreit, Jones, Gomez, Kaiser, and Polosukhin]{vaswani2017attention}
Ashish Vaswani, Noam Shazeer, Niki Parmar, Jakob Uszkoreit, Llion Jones, Aidan~N Gomez, {\L}ukasz Kaiser, and Illia Polosukhin.
\newblock Attention is all you need.
\newblock \emph{Advances in Neural Information Processing Systems}, 30, 2017.

\bibitem[Xu et~al.(2024)Xu, Huang, Hou, Chen, Zhang, Feng, and Xie]{xu2024retrieval}
Jilan Xu, Yifei Huang, Junlin Hou, Guo Chen, Yuejie Zhang, Rui Feng, and Weidi Xie.
\newblock Retrieval-augmented egocentric video captioning.
\newblock In \emph{Proceedings of the IEEE/CVF Conference on Computer Vision and Pattern Recognition (CVPR)}, pages 13525--13536, 2024.

\bibitem[Xue et~al.(2023)Xue, Gao, Xing, Mart{\'\i}n-Mart{\'\i}n, Wu, Xiong, Xu, Niebles, and Savarese]{xue2023ulip}
Le Xue, Mingfei Gao, Chen Xing, Roberto Mart{\'\i}n-Mart{\'\i}n, Jiajun Wu, Caiming Xiong, Ran Xu, Juan~Carlos Niebles, and Silvio Savarese.
\newblock Ulip: Learning a unified representation of language, images, and point clouds for 3d understanding.
\newblock In \emph{Proceedings of the IEEE/CVF Conference on Computer Vision and Pattern Recognition (CVPR)}, pages 1179--1189, 2023.

\bibitem[Yan et~al.(2018)Yan, Shan, and Furukawa]{yan2018ridi}
Hang Yan, Qi Shan, and Yasutaka Furukawa.
\newblock Ridi: Robust imu double integration.
\newblock In \emph{Proceedings of the European Conference on Computer Vision (ECCV)}, pages 621--636, 2018.

\bibitem[Yi et~al.(2023)Yi, Zhou, Habermann, Golyanik, Pan, Theobalt, and Xu]{yi2023egolocate}
Xinyu Yi, Yuxiao Zhou, Marc Habermann, Vladislav Golyanik, Shaohua Pan, Christian Theobalt, and Feng Xu.
\newblock Egolocate: Real-time motion capture, localization, and mapping with sparse body-mounted sensors.
\newblock \emph{ACM Transactions on Graphics (TOG)}, 42\penalty0 (4):\penalty0 1--17, 2023.

\bibitem[Ying et~al.(2007)Ying, Silex, Schnitzer, Leonhardt, and Schiek]{ying2007automatic}
Hong Ying, Carmen Silex, Andreas Schnitzer, Steffen Leonhardt, and Michael Schiek.
\newblock Automatic step detection in the accelerometer signal.
\newblock In \emph{4th International Workshop on Wearable and Implantable Body Sensor Networks (BSN 2007) March 26--28, 2007 RWTH Aachen University, Germany}, pages 80--85. Springer, 2007.

\bibitem[Yonetani et~al.(2024)Yonetani, Baba, and Furukawa]{yonetani2024retailopt}
Ryo Yonetani, Jun Baba, and Yasutaka Furukawa.
\newblock Retailopt: Opt-in, easy-to-deploy trajectory estimation from smartphone motion data and retail facility information.
\newblock In \emph{Proceedings of the 2024 ACM International Symposium on Wearable Computers}, pages 125--132, 2024.

\bibitem[Zeinali et~al.(2024)Zeinali, Zanddizari, and Chang]{zeinali2024imunet}
Behnam Zeinali, Hadi Zanddizari, and Morris~J Chang.
\newblock Imunet: Efficient regression architecture for inertial imu navigation and positioning.
\newblock \emph{IEEE Transactions on Instrumentation and Measurement}, 2024.

\bibitem[Zhang et~al.(2023{\natexlab{a}})Zhang, Gupta, and Zisserman]{zhang2023helping}
Chuhan Zhang, Ankush Gupta, and Andrew Zisserman.
\newblock Helping hands: An object-aware ego-centric video recognition model.
\newblock In \emph{Proceedings of the IEEE/CVF International Conference on Computer Vision (ICCV)}, pages 13901--13912, 2023{\natexlab{a}}.

\bibitem[Zhang et~al.(2023{\natexlab{b}})Zhang, Wang, Li, Huang, Sato, and Lu]{zhang2023structural}
Mingfang Zhang, Jinglu Wang, Xiao Li, Yifei Huang, Yoichi Sato, and Yan Lu.
\newblock Structural multiplane image: Bridging neural view synthesis and 3d reconstruction.
\newblock In \emph{Proceedings of the IEEE/CVF Conference on Computer Vision and Pattern Recognition}, pages 16707--16716, 2023{\natexlab{b}}.

\bibitem[Zhang et~al.(2024)Zhang, Huang, Liu, and Sato]{zhang2024masked}
Mingfang Zhang, Yifei Huang, Ruicong Liu, and Yoichi Sato.
\newblock Masked video and body-worn imu autoencoder for egocentric action recognition.
\newblock In \emph{Proceedings of the European Conference on Computer Vision (ECCV)}, pages 312--330. Springer, 2024.

\bibitem[Zhao et~al.(2023)Zhao, Misra, Kr{\"a}henb{\"u}hl, and Girdhar]{zhao2023learning}
Yue Zhao, Ishan Misra, Philipp Kr{\"a}henb{\"u}hl, and Rohit Girdhar.
\newblock Learning video representations from large language models.
\newblock In \emph{Proceedings of the IEEE/CVF Conference on Computer Vision and Pattern Recognition (CVPR)}, pages 6586--6597, 2023.

\end{thebibliography}
